\documentclass[]{spie}  

\usepackage{times}
\usepackage{amsmath,amsfonts,amssymb}
\usepackage{graphicx}
\usepackage{subcaption}
\usepackage{pgfplots}
\usepackage{booktabs}
\usepackage{float}
\pgfplotsset{compat=1.18}

\usepackage[colorlinks=false]{hyperref}

    \title{Automatic UAV-based Airport Pavement Inspection Using Mixed Real and Virtual Scenarios}

\author{Pablo Alonso}
\author{Jon Ander I\~niguez de Gordoa}
\author{Juan Diego Ortega}
\author{Sara Garc\'ia}
\author{Francisco Javier Iriarte}
\author{Marcos Nieto}
\affil{Fundaci\'on Vicomtech, Basque Research and Technology Alliance (BRTA), Mikeletegi 57, 20009 Donostia-San Sebasti\'an (Spain)}

\authorinfo{Further author information: \\P. Alonso: E-mail: palonso@vicomtech.org}

\begin{document} 
\maketitle

\begin{abstract}
Runway and taxiway pavements are exposed to high stress during their projected lifetime, which inevitably leads to a decrease in their condition over time. To make sure airport pavement condition ensure uninterrupted and resilient operations, it is of utmost importance to monitor their condition and conduct regular inspections. UAV-based inspection is recently gaining importance due to its wide range monitoring capabilities and reduced cost.
In this work, we propose a vision-based approach to automatically identify pavement distress using images captured by UAVs. The proposed method is based on Deep Learning (DL) to segment defects in the image. The DL architecture leverages the low computational capacities of embedded systems in UAVs by using an optimised implementation of EfficientNet feature extraction and Feature Pyramid Network segmentation. To deal with the lack of annotated data for training we have developed a synthetic dataset generation methodology to extend available distress datasets. We demonstrate that the use of a mixed dataset composed of synthetic and real training images yields better results when testing the training models in real application scenarios.   
\end{abstract}

\section{Introduction}
Airport infrastructure maintenance is costly and safety critical. Numerous tasks need to be performed to keep the conditions of airport infrastructure in an excellent state to fulfil the continuous increase in airline traffic.
As there is more demand for airports due to the increase in flight travellers, the load put on taxiways and runways pavement causes 
different types of road distresses. As one of the vital assets of an airport, pavements in good condition do not only guarantee safe operation but also business continuity. These pavement distresses may impact the airport operation capabilities, producing flight cancellations and decreasing airline incomes, but ultimately will reduce airport safety. Therefore, there is a clear requirement to perform continuous inspections of airport facilities. 

Traditional methods to carry out pavement inspection at airports include the use of road vehicles and experienced personnel to visually identify road distresses such as cracks or holes. These methods allow to identify all the defects present in the inspection area, however, it requires the inspection area to be closed to airplane traffic and usually requires several hours to complete the full mission. 

The advances in Unmanned Aerial Vehicle (UAV) technologies have allowed the application of affordable and compact unmanned aircrafts in different situations. The research community has increasingly supported the use of UAV to perform infrastructure inspection in different domains \cite{Irizarry2012drone, seo2018drone}, including airport inspections \cite{hubbard2018uas}. Using UAVs to perform visual inspection has attracted considerable attention from researchers and transportation agencies. Besides, equipping the UAV with a camera expands the possibilities of using it with Computer Vision (CV) methods for automatic assessment of the quantity and severity of distress based on visually identifiable image features. 

The development of CV methods has benefited from the advances in Machine Learning (ML) and Deep Learning (DL) technologies. The DL methods used to assess and identify pavement distress are based on supervised learning approaches which require an annotated training dataset to generate the inference models. These ML models can then be integrated and embedded in the UAV onboard processing unit or used on an on-ground station. Onboard processing avoids the necessity of streaming video to the ground and thus reducing the bandwidth requirements. However, the computing power is limited and models have to be designed accordingly. Moreover, to generate robust and reliable detection models, these DL methods require large volumes of annotated image datasets which are not easy to collect. 

There are several challenges when collecting image-based pavement datasets. First, organizational and logistic coordination with airport authorities could make the data capturing process slower or even not possible due to airport operations. Second, preparing the data recording process requires efforts to technically process, synchronize, transform, and compress large volumes of raw data. Third, labelling a large volume of images is a very time-consuming task which requires to have many well-instructed human annotators. In addition, learning methods for pavement distress inspection usually require to have pixel-level mask labels for the training process. Such types of tasks are one of the most complex and time-consuming annotation tasks in CV. Moreover, it is prone to human error for many reasons: low-resolution images, bad lighting conditions, fuzzy transitions between distress and background, intrusive objects, non-constant distress width, inappropriate labelling tools, etc. Because of this, manual annotations tend to be inaccurate, particularly at pixel level. These inaccuracies introduced by human annotators could bias the training process compromising the prediction capabilities of the model.   

Therefore, to overcome the previous difficulties this work proposes a methodology to generate synthetic image sequences using a hyperrealistic UAV virtual simulator. The resulting annotated virtual sequences are combined with public crack detection datasets, to train state-of-the-art pavement distress segmentation DL models. To achieve this objective, we have designed and modelled a close-to-real airport virtual scenario using a modern graphics engine, including realistic road features, different instances of defects and different lighting and appearance conditions such as diverse weather effects. We propose to use a mixed dataset of synthetic and real annotated images to train a segmentation DL network based on EfficientNet feature extraction and Feature Pyramid Network (FPN) as segmenter, which we aim to run in an embedded system mounted in the drone. Due to the lack of public datasets including multiple types of defects, we have focused our experiments on detecting cracks only. To tackle class imbalance in the dataset the Dice Loss optimization is applied when training the models. 

\section{Related work}
Image segmentation is the process of dividing the image into multiple segments for easier analysis. There are multiple methods of performing the segmentation, from image processing techniques to Artificial Neural Networks \cite{khanSurvey2013}. Segmentation techniques have been used in multiple domains \cite{kaurWatershed2014, thannamalHistogram2014, richardFuzzy2013}. In this work, we will focus on the usage of these techniques in pavement distress detection.
\subsection{Pavement distress detection}
In recent years, many ML and Convolutional Neural Network (CNN) approaches have been proposed for automatically detecting cracks and other pavement distress. Zhang et al. \cite{zhang2016road}, used a CNN to segment road cracks using image patches of a self-collected dataset. They achieved better performance than previous ML methods such as Support Vector Machines or boosting methods. Jenkins et al. \cite{jenkins2018deep} applied the U-Net architecture for pixel-wise segmentation of road and pavement surface cracks, using the public CrackForest dataset \cite{shi2016automatic}. In \cite{Zou2019deepcrack}, DeepCrack was proposed, a crack detection method using hierarchical convolutional features. This method combines features at different levels in a U-Net-like architecture to produce crack segmentation masks.

Yang et al. \cite{yang2019feature} introduced a feature pyramid network for segmentation of pavement cracks. In this method, they use an encoder to extract features from the image (top-down pathway), and a bottom-up pathway which produces features at different scales. These features are put together using a boosting mechanism and processed to obtain the final segmentation mask. This method was proven to outperform previous methods in various datasets and to have better generalizability.

These crack segmentation methods in the literature focus on the model itself and how it performs given various datasets. They don't discuss their applications and don't validate their findings in real-world scenarios, which is a topic mostly unexplored in this domain.

Alternatively, other methods resort to the use of methods based on object detection such as the work by Du et al. \cite{Du2020PavementDD}, using the popular YOLO network, where the resulting output of the trained model is a list of bounding boxes of the detected pavement distress. However, when interleaving cracks appear in the image, it is difficult to identify each individual crack and its boundaries.

For addressing the class imbalance, Dice Loss was introduced by Milletari et al. \cite{Milletari2016dice} for medical image segmentation. This loss function leverages the Dice coefficient and does not need to establish pre-computed class weights for achieving good performance. This technique has also been used in some recent works for pavement crack segmentation \cite{Wang2020pan}.


\subsection{Synthetic dataset applications}
Manual annotation of real images for Deep Learning approaches is time-consuming and expensive, especially for segmentation tasks that require a pixel-level annotation. Because of this, the generation of synthetic datasets with automated ground truth generation has been proposed in multiple approaches in order to replace or complement real images. Rill-Garc\'ia et al. \cite{visapp22} introduced the Syncrack tool, which allows users to generate synthetic annotated images of pavement and concrete textures, but the models trained with Syncrack data showed a considerably lower recall than those trained on real datasets. In \cite{Kanaeva_2021}, synthetic data augmentation was obtained by overlapping crack images from the CrackForest dataset \cite{shi2016automatic} into the roads in the KITTI and Cityscapes dataset. The models trained on such data, however, were sensitive to shadows and light conditions. In \cite{pbgms}, 3D synthetic environments with Physics-based Graphic Models (PBGMs) were employed for the labelling and validation of structural inspections using UAVs. 

\section{Synthetic dataset generation} \label{sec:datasetGen}
In order to obtain a greater data variety to train and validate our model, we generated a synthetic dataset of pavement defects using the AirSim simulator \cite{airsim2017fsr} on a custom Unreal Engine environment \cite{unrealengine}.

\subsection{Environment design}
\begin{figure}
    \centering
    \includegraphics[width=0.9\textwidth]{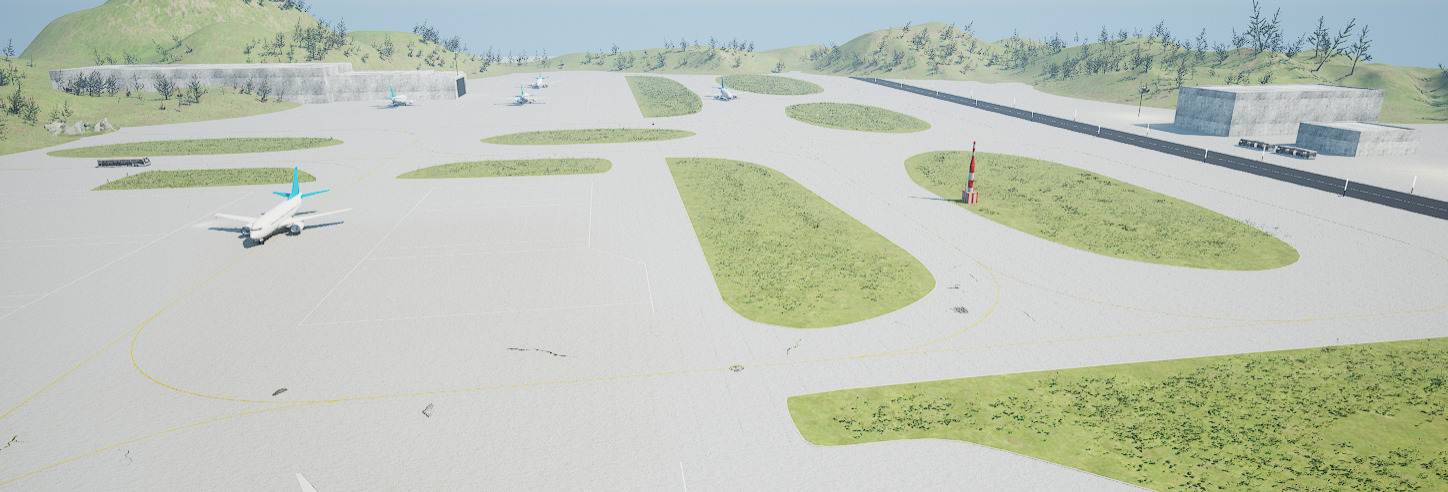}
    \caption{Designed virtual airport environment.}
    \label{fig:airport}
\end{figure}

The first step in the pavement defect dataset generation was to create the environment in which the dataset recordings would take place. This environment was designed using Unreal Engine 5, which offers more photo-realistic lighting than previous generations. The designed environment consists of a custom airport with its runway, taxiways, hangers, apron and terminals. The buildings were created using 3D modelling software and 
the runway and taxiways are built of concrete tiles with a size of $5 \times 5$ meters. The tile dimensions were chosen based on the approximated size of real airport tiles. A screenshot of the virtual airport is shown in Figure \ref{fig:airport}.

\begin{figure}[t]
    \centering
    \includegraphics[width=0.85\textwidth]{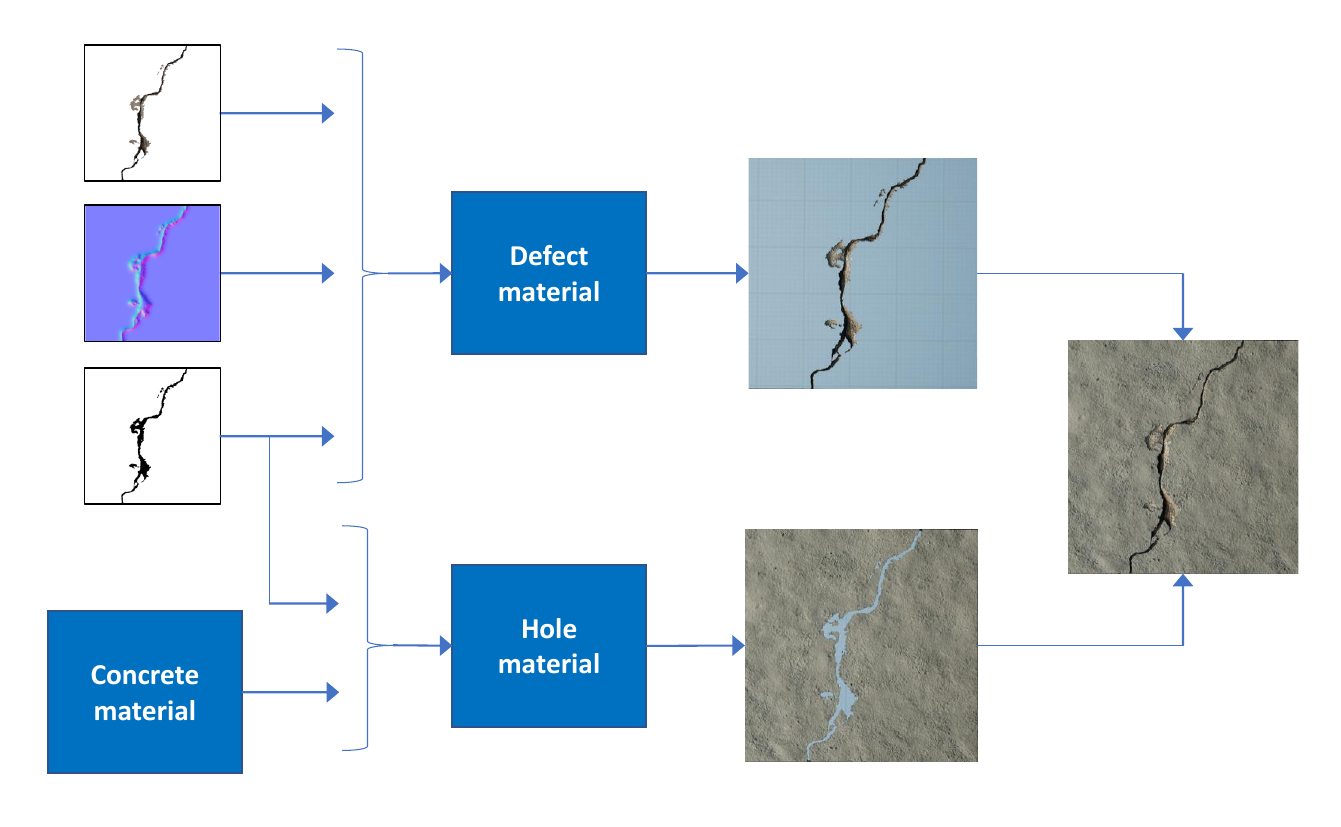}
    \caption{Synthetic pavement defect generation process.}
    \label{fig:defect_design}
\end{figure}

Next, the pavement defects are introduced into the airport environment. The steps to design and spawn pavement defects are shown in Figure \ref{fig:defect_design}. First, the textures required to create the concrete defects were generated using image editing software. 
The textures for each defect consist of an RGB mask based on a real image of a pavement defect, the normal map of the defect to describe its geometry, which is created using the gradients of the RGB texture, and a binary opacity mask that delimits the boundaries of each defect. Using these textures, two materials were created for each defect in order to ease the dataset annotation process. On the one hand, a material containing the defect itself was generated. On the other hand, a new material was created for the pavement tile that would hold the defect. This new tile presents a hole right where the defect is located so that, when the defect and the tile are spawned in the environment, the defect fits exactly in the hole. 
The holed tile material was generated using the original tile material and applying the negative binary mask of its corresponding defect.

Once the defect materials were generated, a static mesh was generated for each defect and its holding tile. These meshes were spawned along the runway and taxiway of the airport, replacing some original tiles in our Unreal Engine environment. Note that in order to generate long cracks (i.e. cracks longer than the size of each tile), several defective tiles were spawned in a row, so that the same crack would span multiple tiles.

\subsection{Data capture}
The synthetic dataset was generated employing the AirSim simulator on the created airport environment. AirSim is an open-source simulator for drones and other vehicles which supports software-in-the-loop~(SIL) simulation that enables the design of multi-sensor and multi-modal scenarios. It works as an Unreal Engine plugin, and offers an API that allows users to control the vehicle and retrieve physically and visually realistic data from the environment, using the virtual sensors.

In the simulations executed with AirSim, a multirotor drone was spawned into the airport environment, and hovered along the runway and taxiways. The spawned drone was equipped with two cameras (an RGB and a segmentation camera), both aiming downwards, that is, to the floor. The RGB camera retrieves colour images from the environment, given its previously defined intrinsic parameters and characteristics such as noise or motion blur. The segmentation camera generates the ground truth annotation for the colour image: it specifies the class of each pixel in the RGB image (defect or no-defect). When a simulation starts, every mesh in the airport environment is first given the same object ID employing the AirSim API, so that every object is seen as black by the segmentation camera. Then, since the name of the static meshes for each defect is known (it can be checked in the Unreal Editor), another object ID is given to the defect meshes. This way, the segmentation camera retrieves every pixel as black, except for those that actually belong to a defect.

\begin{figure}
    \centering
    \begin{subfigure}{.31\textwidth}
    \centering
    \includegraphics[width=.9\textwidth]{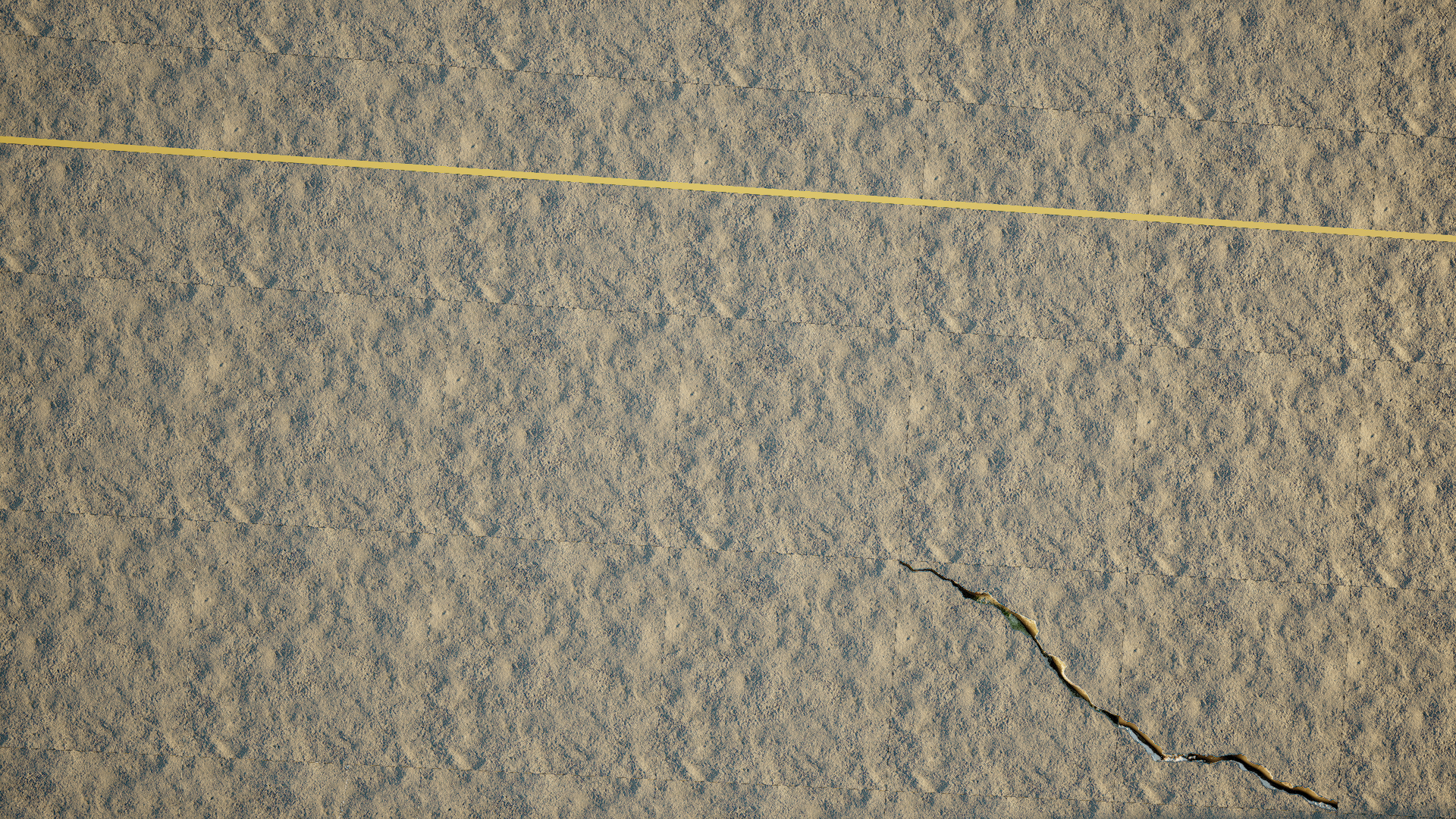}
    \end{subfigure}%
    \begin{subfigure}{.31\textwidth}
    \centering
    \includegraphics[width=.9\textwidth]{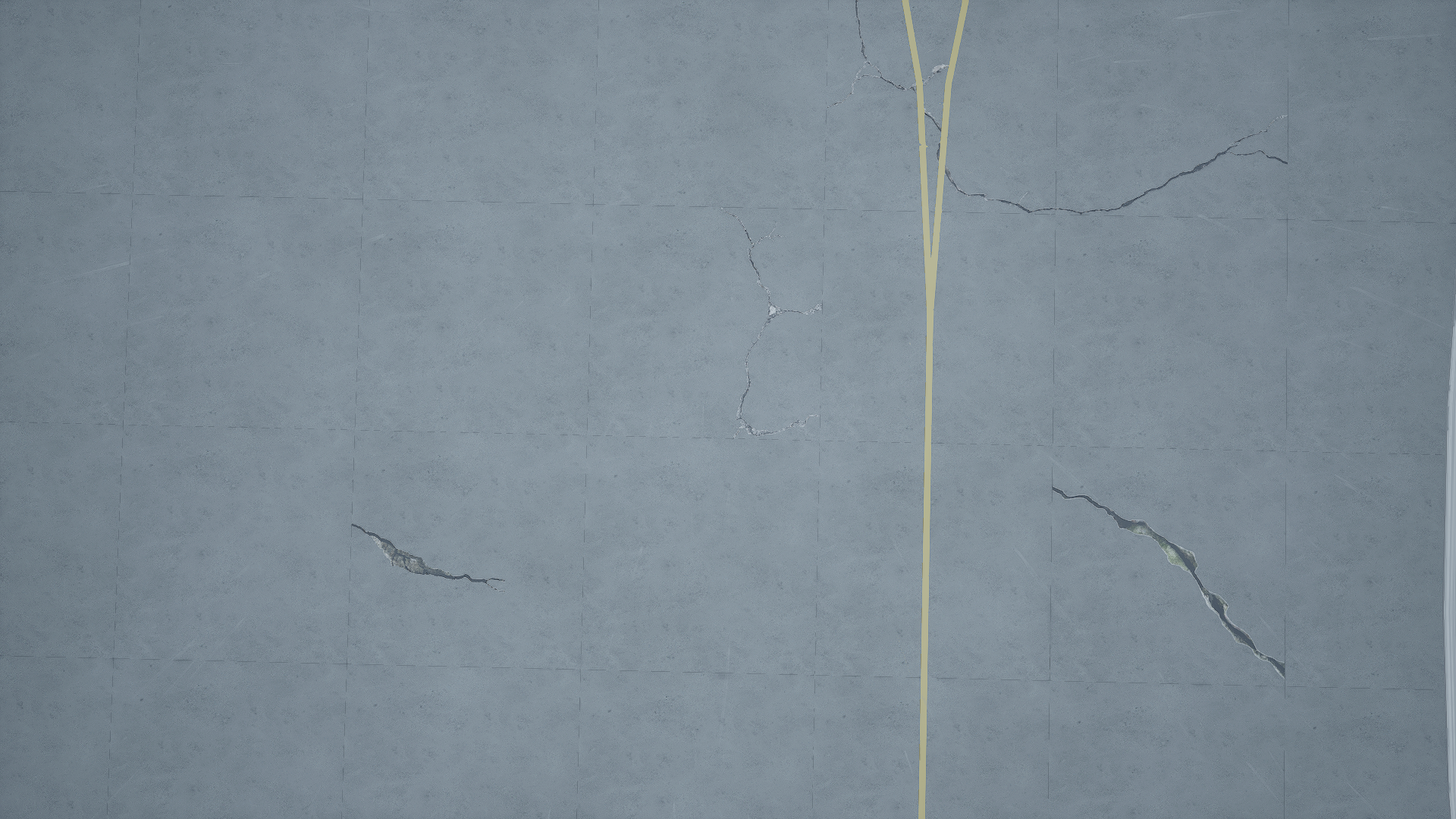}
    \end{subfigure}
    \begin{subfigure}{.31\textwidth}
    \centering
    \includegraphics[width=.9\textwidth]{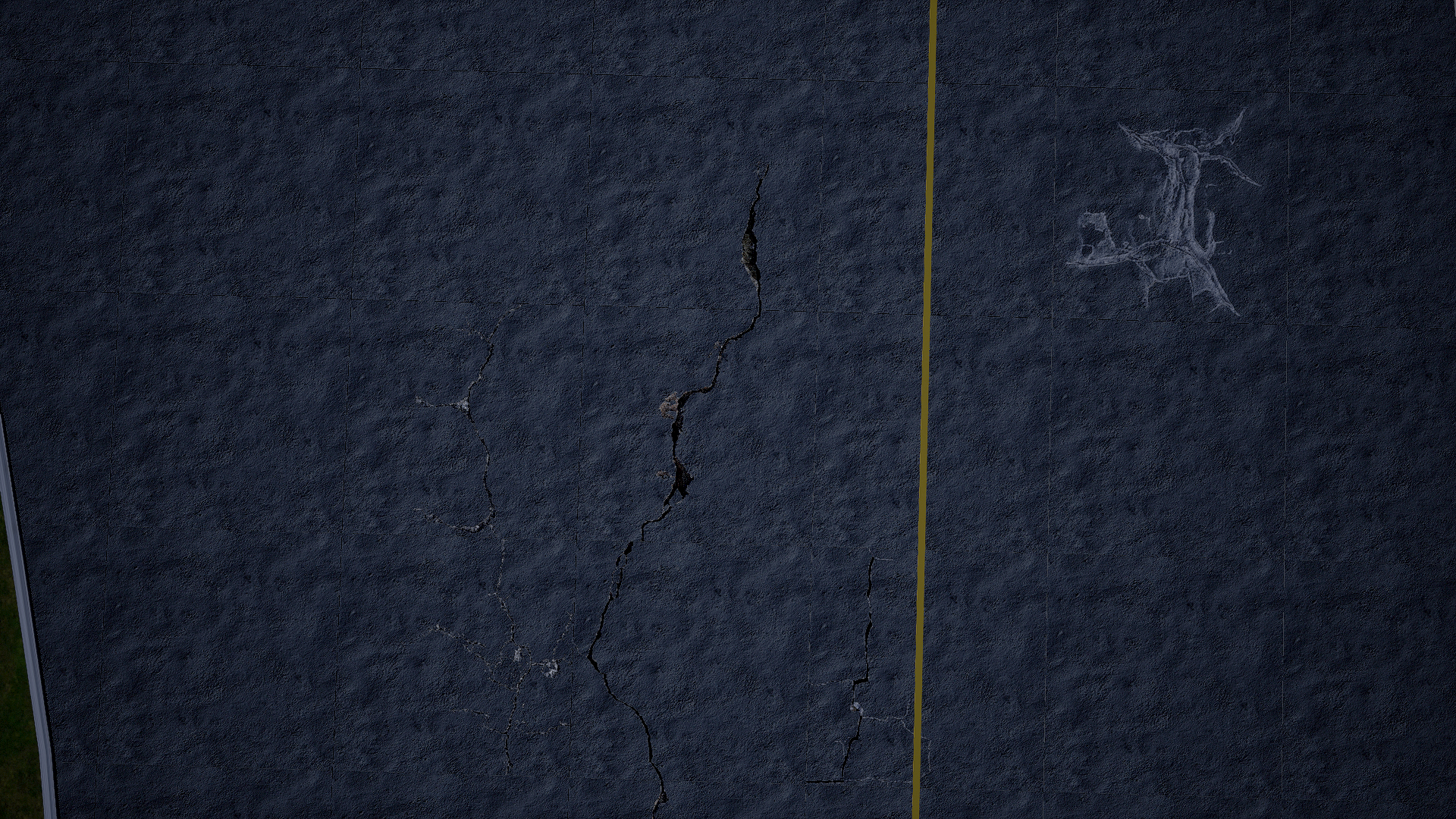}
    \end{subfigure}%

    \vspace{1ex}
    
    \begin{subfigure}{.31\textwidth}
    \centering
    \includegraphics[width=.9\textwidth]{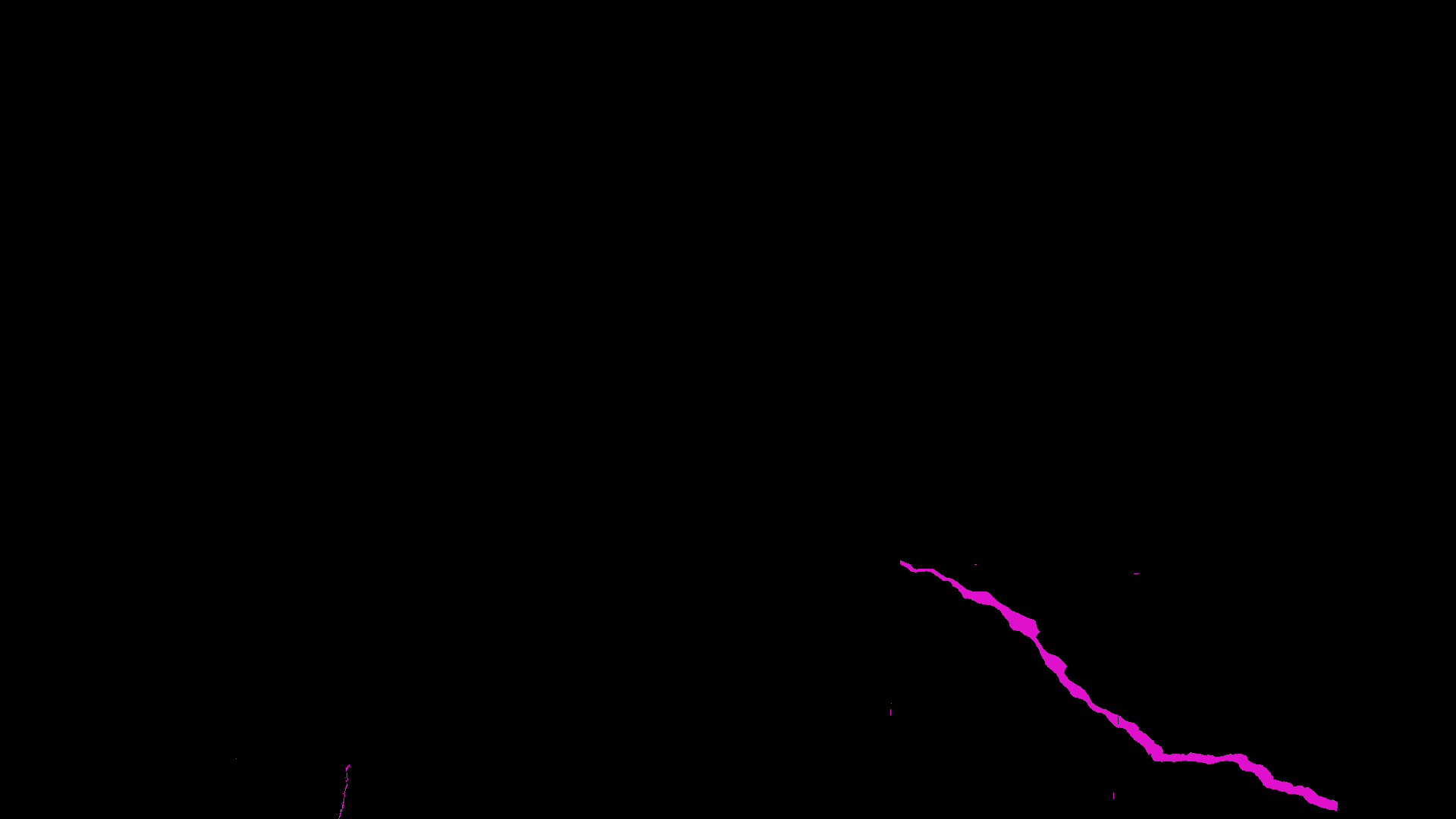}
    \caption{}
    \label{fig:syn_dataa}
    \end{subfigure}%
    \begin{subfigure}{.31\textwidth}
    \centering
    \includegraphics[width=.9\textwidth]{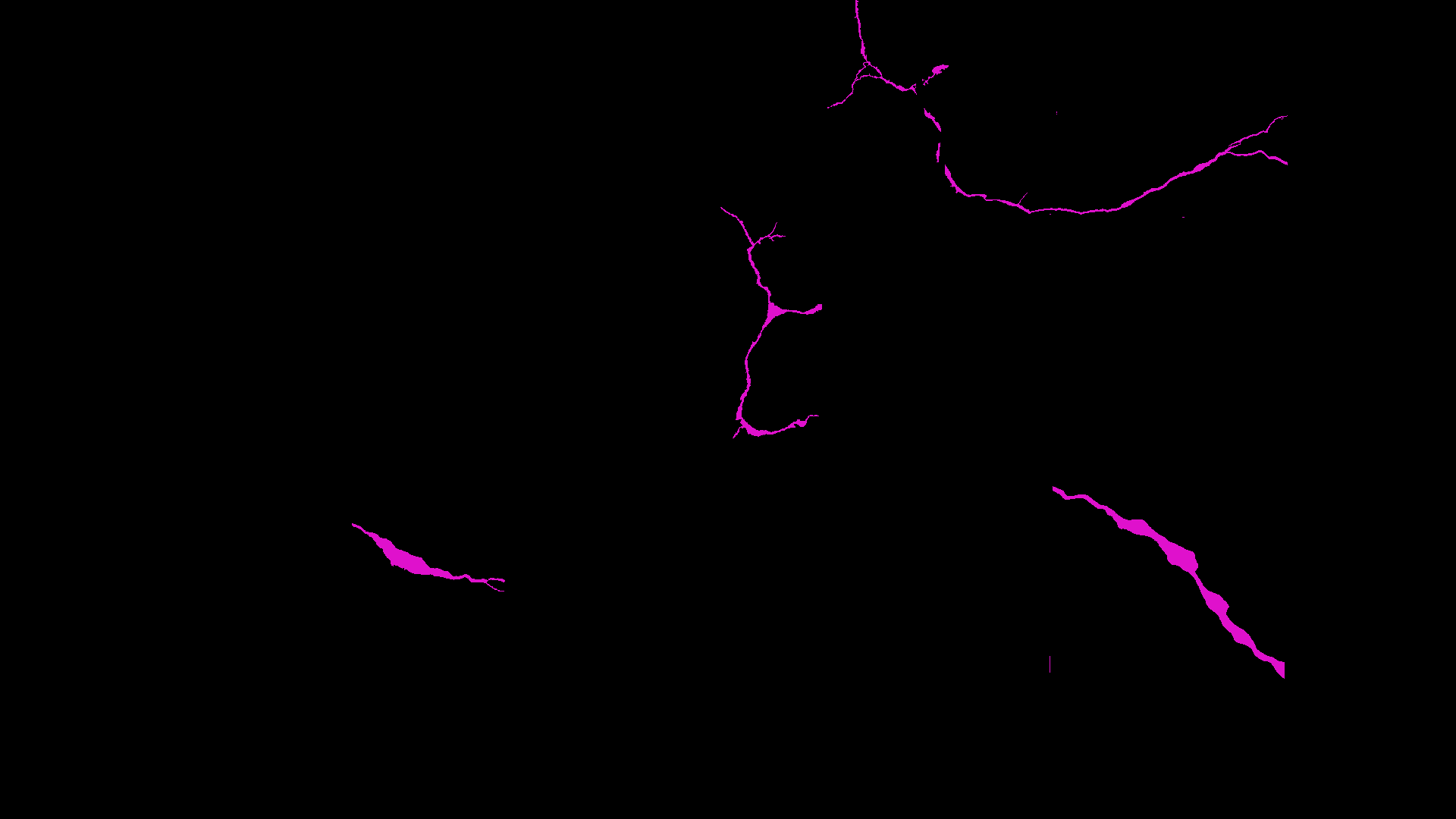}
    \caption{}
    \label{fig:syn_datab}
    \end{subfigure}
    \begin{subfigure}{.31\textwidth}
    \centering
    \includegraphics[width=.9\textwidth]{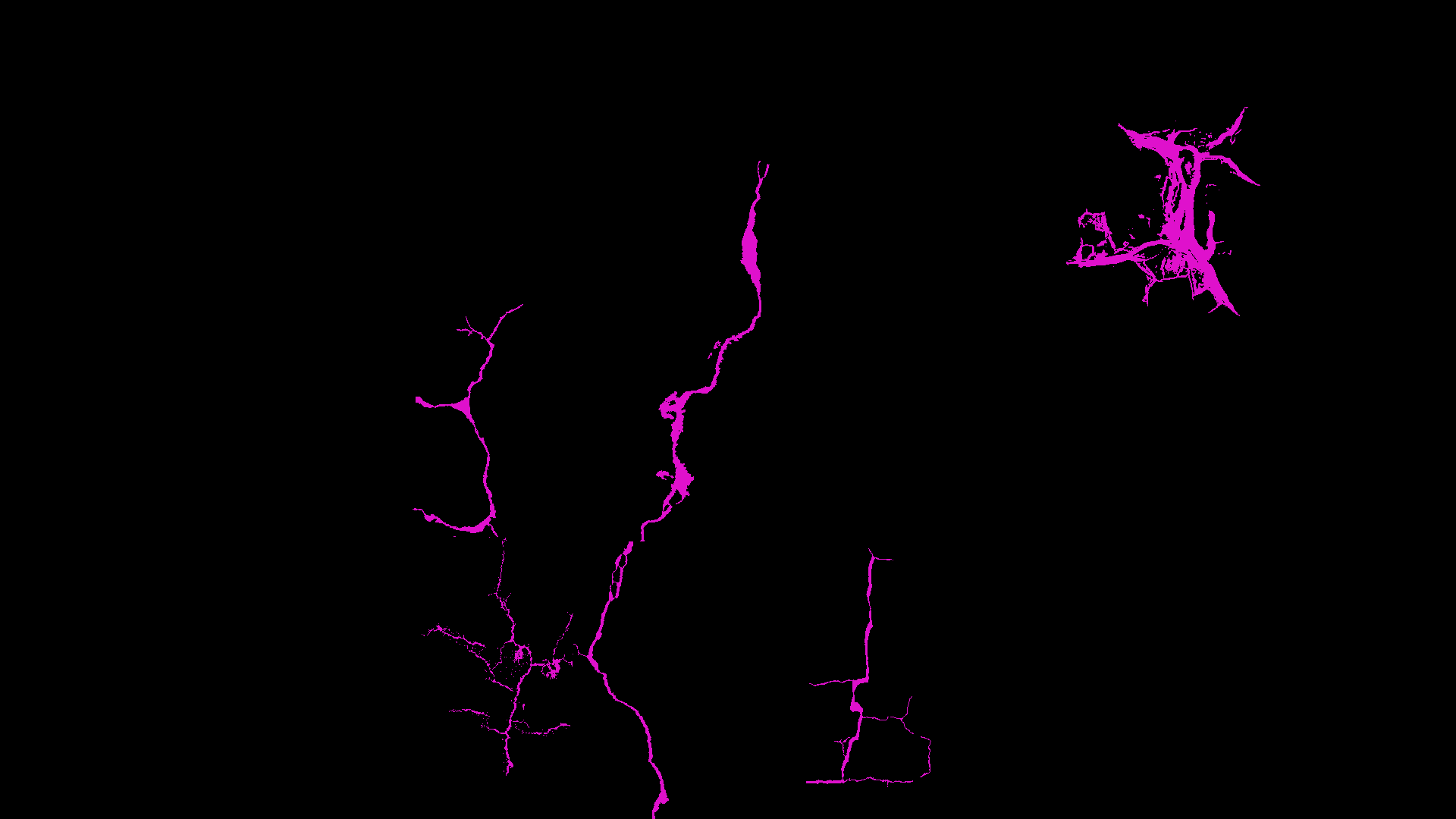}
    \caption{}
    \label{fig:syn_datac}
    \end{subfigure}%

    \vspace{1ex}
    \caption{RGB and corresponding annotation samples from the generated synthetic dataset. Samples (\subref{fig:syn_dataa}), (\subref{fig:syn_datab}) and (\subref{fig:syn_datac}) correspond to simulations at dusk, noon rain and at night, respectively.}
    \label{fig:syn_data}
\end{figure}

In order to generate a robust dataset, it is necessary to simulate the environment under different lighting conditions. For the generation of our synthetic dataset, the Ultra Dynamic Sky package was employed \cite{UltraDynamicSky}, which allows to produce simulations at multiple times of the day and in different weather conditions (rainy, cloudy, clear skies, fog, etc.). Moreover, the synthetic environment is rounded off by introducing runway lights poles and spotlights into the airport, which illuminate the scene at night. The AirSim simulator was run in multiple atmospheric conditions, and the retrieved RGB images and annotations in each simulation were saved in order to generate our synthetic dataset. Figure \ref{fig:syn_data} shows samples from the synthetic dataset that were generated under different atmospheric conditions.

\section{Pavement defect detection methodology}
\subsection{Network architecture}

\begin{figure}[t]
    \centering
    \includegraphics[angle=90,origin=c,width=\textwidth,trim=160 10 0 0]{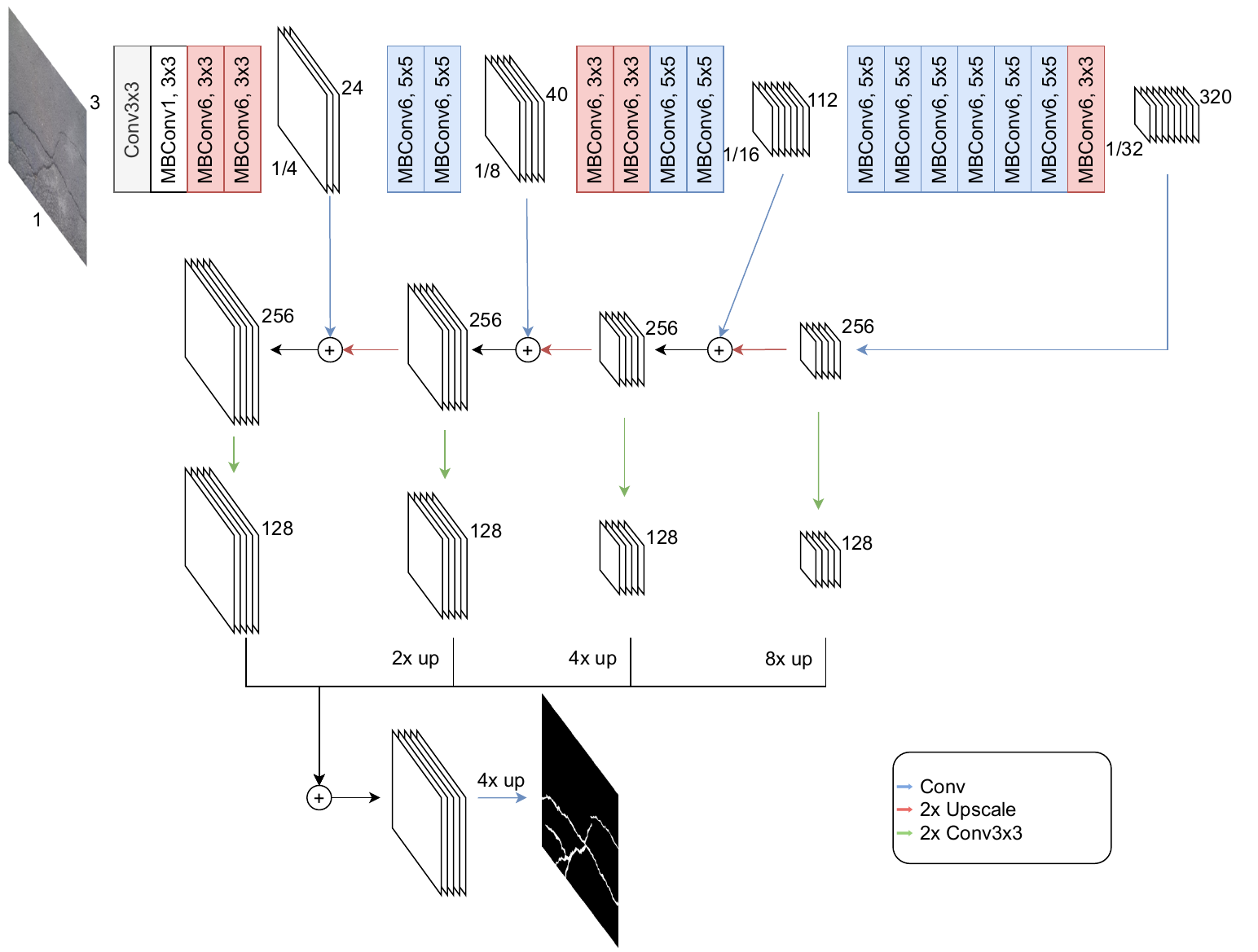}
    \caption{EfficientNet-FPN segmentation network architecture.}
    \label{fig:architecture}
\end{figure}

Our network is composed of a feature-extracting backbone and a segmentation head. For the backbone, we used the B1 variant of EfficientNet \cite{tan2019efficientnet}, pre-trained on ImageNet for classification and performed transfer learning. EfficientNet proposes a good balance between accuracy and performance and went in line with our performance constraints on an embedded platform. We also need to input a high-resolution image to the network, as the defects in the image are only a few pixels wide when captured from the height of a drone and small details can be lost when downscaling the image.

Similarly to Yang et al.\cite{yang2019feature}, we used a Feature Pyramid Network (FPN) as our segmenter \cite{lin2017fpn, kirillov2017unified}. From our experiments, it retains enough high-frequency information needed for segmenting finer cracks and defects but is not as computationally expensive as other popular networks such as U-Net, which perform more computation at a high resolution. This network combines the features from the encoder at different scales using a pyramid structure, segmenting the image at different levels and assembling it to combine high- and low-resolution details, with semantically strong and low-level features. In contrast with other, also popular methods for crack segmentation, such as U-Net, the model outputs a segmentation mask of a lower resolution than that of the input image, which is then upscaled as a final step to match the original input.

The full architecture of our network is displayed in Figure~\ref{fig:architecture}.

One big challenge inherent to this task is the class imbalance that exists in the data. The background pixels are more prominent than the crack pixels (see Table \ref{tab:datasets}), so using the popular ``binary cross-entropy'' loss function is not efficient. Having the whole image classified as non-crack yields a good overall score, as most of the image is indeed non-crack. Consequently, the network tends to converge towards that (the so-called ``All-Black'' issue). To tackle this problem, we used a combination of Generalised Dice Loss \cite{Sudre2017dice} and binary cross-entropy. Dice loss optimises the S\o{}rensen-Dice coefficient, or F1-score, which deals with class-unbalanced tasks more efficiently.

\subsection{Data augmentation}
In addition to the synthetic data generation method described in Section~\ref{sec:datasetGen}, we perform further geometric and image transformations for every input.

For geometric transformations, we randomly flip, rotate and scale the image. We also perform an elastic transform as described by Simard et al. \cite{simard2003bestpractices}, which applies displacement fields in order to "deform" the shape of the objects in the image. This way, we greatly enhance the variety of the cracks present in our dataset. This is especially relevant in the synthetically generated images, as we manually modelled a finite number of realistic defects, and the network might overfit those samples.

For image transformations, we apply random brightness, contrast, gamma and hue shifts, as well as sharpening and blurring with randomised parameters.

We also applied an artificial motion blur augmentation to the public crack datasets. Most of these images were captured with a static camera, so the images are sharp. Capturing video from a moving drone will result in blurred frames due to motion, especially in lower light conditions where a slower shutter speed is used. Therefore we use motion blur data augmentation to try and simulate that effect in the training set.

\section{Experiments}
\subsection{Implementation details}
Performing automatic visual inspection using a UAV can be achieved by defining different operational use cases. One option is to capture the required images and transferred them to a ground processing station. As an alternative, the image analysis algorithm can be embedded in a UAV image processing unit. In this work, we explore the latter by designing our network taking into account compatibility with embedded systems such as Nvidia Jetson AGX Xavier devices. Having the computation onboard instead of off-site enables us to reduce the connectivity and bandwidth requirements of the drone, as data is only recorded when a defect is detected in the pavement. However, it implies that computational efficient DL networks are needed to guarantee real-time performance while maintaining a good level of detection accuracy.  

The DL network training was performed using PyTorch and the Segmentation Models Pytorch library \cite{yakubovskiy2019smp}, which enables easy access to multiple backbones and segmentation models. The backbone was pre-trained for classification with ImageNet, performing transfer learning for feature extraction. All models were trained for 100 epochs using the One Cycle learning rate scheduler \cite{smith2017superconvergence}. We evaluate the training process after every epoch and kept the weights that produced the highest Intersection over Union (IoU).

For inference, getting the most performance out of the hardware was essential, so we optimised our model using the TensorRT framework, it is optimized to achieve good performance on Jetson devices. The model was exported from \mbox{PyTorch} to ONNX before performing inference on \mbox{TensorRT} using a ROS node written in C++. The performance metrics of our model with various resolutions are shown in Table~\ref{tab:trt_performance}. In our use case, $1920\times1088$ will be used as input resolution.

\begin{table}
\begin{center}
\begin{tabular}{ l l l }
    \toprule
    \textbf{Input resolution} & \textbf{FPS} & \textbf{Latency}\\
    \midrule
    $1920\times1088$ & $9.01$ & $111$ ms \\
    $1280\times736$ & $19.01$ & $53$ ms \\
    $1600\times928$ & $12.31$ & $81$ ms \\
    \bottomrule
\end{tabular}
\end{center}
\caption{Performance metrics for different input resolutions on our EfficientNet-FPN network optimised with TensorRT. The data type is FP16 and only network inference time is measured.}
\label{tab:trt_performance}
\end{table}

\subsection{Datasets}
We combined multiple public crack datasets for the training and validation sets to obtain the greatest variety in style, shape and scale. As explained in Section~\ref{sec:datasetGen}, we also add the synthetic datasets. The datasets we used and their details are presented in Table~\ref{tab:datasets}.

\begin{table}[t]
\begin{center}
\begin{tabular}{l r r l l}
    \toprule
    \textbf{Dataset name} & \textbf{Training \#} & \textbf{Validation \#} & \textbf{Resolution} & \textbf{\% of crack pixels} \\
    \midrule
    CRACK500 \cite{yang2019feature, zhang2016road} & $1896$ & $348$ & $640\times360$ & $6.01\%$ \\ 
    GAPs384 \cite{eisenbach2017how, yang2019feature} & $465$ & $44$ & $540\times\{440,640\}$ & $1.19\%$ \\
    CrackForest \cite{shi2016automatic} & $90$ & $18$ & $480\times320$ & $2.28\%$ \\
    DeepCrack 
    \cite{liu2019deepcrack} & $300$ & $237$ & $544\times384$ & $3.53\%$ \\
    Synthetic dataset V1 & $2293$ & $574$ & $1920\times1080$ & $1.04\%$ \\ 
    Synthetic dataset V2 & $406$ & $102$ & $1920\times1080$ & $0.46\%$ \\
    \bottomrule
\end{tabular}
\end{center}
\caption{Training and validation dataset metadata. Note that we used the cropped version of the CRACK500 dataset, where images are divided into multiple lower-resolution images.}
\label{tab:datasets}
\end{table}

As all datasets have different image resolutions, we perform training on random 320x320 crops of the original image. The cropping is performed at training time as a preprocessing step, so the data introduced to the network is different in each epoch. For validation and evaluation, we input the full image to the network (with padding if needed) and a batch size of 1.

\section{Results and discussion}
In this section, we evaluate the results of networks trained on combinations of synthetic and real images. Evaluation is done using the available, annotated public datasets, as well as a small test dataset from synthetic drone capture of runways.

For comparison between models, we used the metrics Precision, Recall and F1-Score. The confidence threshold in order to consider an output from the sigmoid function of the network to be positive is $0.5$.

\subsection{Results on public crack datasets}
We used the annotations in the validation or test sets from the public datasets to evaluate the performance. In cases where the authors didn't divide the data into those splits or in our generated synthetic datasets, we randomly chose 80\% of the images for use in training, and the remaining 20\% for validation and testing (see Table~\ref{tab:datasets}). The metrics are calculated throughout all datasets, which are combined to form a large test set.

The metrics of our experiments are presented in Table~\ref{tab:publicmetrics}.

\begin{table}
\begin{center}
\begin{tabular}{r r r r r r}
    \toprule
    \textbf{ID} & \textbf{Real \#} & \textbf{Synth. \#} & \textbf{Precision} & \textbf{Recall} & \textbf{F1-score} \\
    \midrule
    1 & $2915$ & $0$ & $0.7921$ & $0.7794$ & $\mathbf{0.7556}$ \\ 
    2 & $0$ & $2699$ & $\mathbf{0.8213}$ & $0.2806$ & $0.2493$ \\
    3 & $2915$ & $2699$ & $0.7510$ & $\mathbf{0.8097}$ & $0.7441$ \\ 
    \bottomrule
\end{tabular}
\end{center}
\caption{Evaluation metrics of our EfficientNet-FPN network in a combined test dataset from multiple public datasets. In each experiment, the network was trained with different data, while all other parameters remained the same.}
\label{tab:publicmetrics}
\end{table}

In the first experiment, where only images from public datasets were used, we obtained the best results among all experiments. When introducing our synthetic dataset for training along with the real images, the F1-score decreased, but the Recall value increased. This means that the network can detect more cracks in the image, at the expense of more false positives. This phenomenon can also occur when lowering the threshold for determining if an output from the network is positive or negative. Hence, in the public crack datasets, there is not significant variation in the results between using a mixed dataset with synthetic and real data or only real images. Introducing extra synthetic data from another domain has little effect when the images from the test set and the training set of the public datasets are similar.

When training is performed using only synthetic images, the network cannot generalise and the performance is low. Precision is higher than the others, but recall is low, which indicates that most of the defects in the image are not detected.



In Table~\ref{tab:sota_comparison} we compare our model with other state-of-the-art methods for crack segmentation on the CRACK500 dataset \cite{yang2019feature, zhang2016road}. The metrics chosen are widely used to evaluate segmentation tasks. We use the Intersection over Union (IoU); Optimal Dataset Scale (ODS) and Optimal Image Scale (OIS) \cite{arbelaez2009contours}, which choose the threshold which yields the best F1-score on a dataset level and an image level respectively; and the F1 score at a fixed threshold of $0.5$.

We improve over the Feature Pyramid by Yang et. al. \cite{yang2019feature}, who use a similar network architecture to ours. Despite not reaching the performance level achieved by other state-of-the-art methods such as the U-Net and the method by Wang et. al., we are constrained by the computational power of an embedded platform, which has to perform inference in real-time. Thus, we cannot leverage the performance benefits a more complex model can provide. 

\begin{table}
    \centering
    \begin{tabular}{r l l l l}
        \toprule
        \textbf{Model} & \textbf{IoU} & \textbf{ODS} & \textbf{OIS} & \textbf{F1} \\
        \midrule
        DC-DSN \cite{li2021pavement} & -- & $0.627$ & $0.669$ & -- \\
        FPHBN \cite{yang2019feature} & $0.489$ & $0.604$ & $0.635$ & -- \\
        PAN \cite{Wang2020pan} & $0.624$ & -- & -- & $0.79$ \\
        U-Net \cite{Lau2020AutomatedPC} & -- & -- & -- & 0.733 \\
        Wang et. al. \cite{Wang2021convolutional} & \boldmath$0.737$ & -- & -- & \boldmath$0.827$ \\
        Ours & $0.56$ & \boldmath$0.704$ & \boldmath$0.716$ & $0.701$ \\
        \bottomrule
    \end{tabular}
    \caption{Performance metric comparison between various SOTA crack segmentation models on the CRACK500 dataset.}
    \label{tab:sota_comparison}
\end{table}

\subsection{Results on airport runway recordings}
In our research, we didn't find any public recordings of airport runway inspections that had annotated cracks or other defects. Therefore, we annotated a few images from 2 different recordings of drone capture of airport runways or taxiways.

Due to the nature of the defects to be detected and the quality of the recordings, we cannot guarantee a high precision of annotation, as the footage is blurry at times due to the motion of the camera, lighting conditions, etc., and the boundaries of the defects are unclear. However, these images are taken directly from our target domain, therefore are more valuable for our use case than public dataset images, and rough annotations should serve as a point of comparison between models.

\begin{figure}
    \centering
    \begin{subfigure}{.2\textwidth}
    \centering
    \includegraphics[width=.9\textwidth]{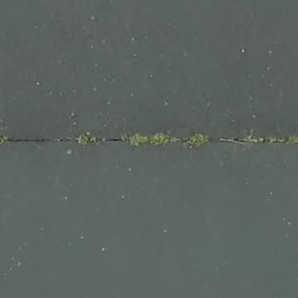}
    \caption{}
    \label{fig:dataexamplea}
    \end{subfigure}%
    \begin{subfigure}{.2\textwidth}
    \centering
    \includegraphics[width=.9\textwidth]{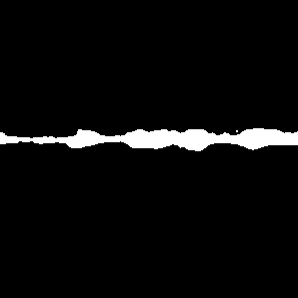}
    \caption{}
    \label{fig:dataexampleb}
    \end{subfigure}
    \begin{subfigure}{.2\textwidth}
    \centering
    \includegraphics[width=.9\textwidth]{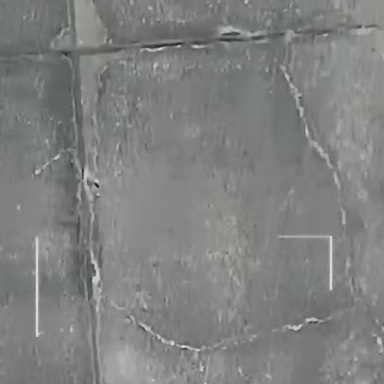}
    \caption{}
    \label{fig:dataexamplec}
    \end{subfigure}%
    \begin{subfigure}{.2\textwidth}
    \centering
    \includegraphics[width=.9\textwidth]{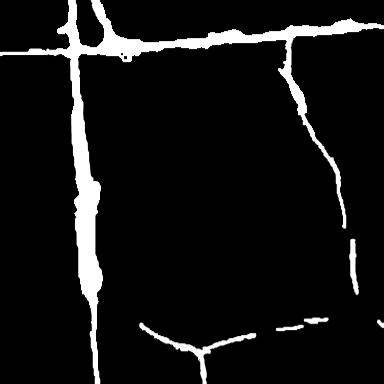}
    \caption{}
    \label{fig:dataexampled}
    \end{subfigure}
    \caption{Detail of the annotation from the airport captures. (\subref{fig:dataexamplea}) and (\subref{fig:dataexamplec}) are the images, and (\subref{fig:dataexampleb}) and (\subref{fig:dataexampled}) are the annotations.}
    \label{fig:dataexample}
\end{figure}

Some examples of our annotations are presented in Figure~\ref{fig:dataexample}.

We perform a similar evaluation in this validation dataset which was not used for training the models. The evaluation metrics of our experiments are presented in Table~\ref{tab:runwaymetrics}.

\begin{table}
\begin{center}
\begin{tabular}{r r r r r r}
    \toprule
    \textbf{ID} & \textbf{Real \#} & \textbf{Synth. \#} & \textbf{Precision} & \textbf{Recall} & \textbf{F1-score} \\
    \midrule
    4 & $2915$ & $0$ & \boldmath$0.4406$ & $0.0845$ & $0.1367$ \\ 
    5 & $0$ & $2699$ & $0.3527$ & $0.2534$ & $0.2574$ \\
    6 & $2915$ & $2699$ & $0.3391$ & \boldmath$0.3823$ & \boldmath$0.3457$ \\ 
    \bottomrule
\end{tabular}
\end{center}
\caption{Evaluation metrics of our EfficientNet-FPN network in a self-annotated dataset of airport runway capture. In each experiment, the network was trained with different data, while all other parameters remained the same.}
\label{tab:runwaymetrics}
\end{table}

When training with just the real data from public datasets, the network isn't able to generalise to runway recordings, which are generally captured from a higher altitude and the appearance of the defects is different. In this domain, training with just synthetic images yields better results than just real images. 
Overall, the best performance is achieved using the mixed dataset.

\begin{figure}
    \centering
    \begin{subfigure}{.16\textwidth}
    \centering
    \includegraphics[width=.95\textwidth]{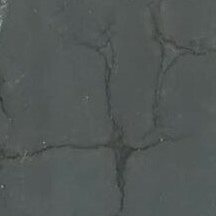}
    \end{subfigure}%
    \begin{subfigure}{.16\textwidth}
    \centering
    \includegraphics[width=.95\textwidth]{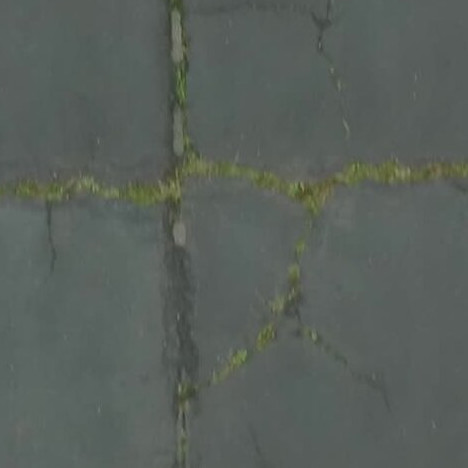}
    \end{subfigure}%
    \begin{subfigure}{.16\textwidth}
    \centering
    \includegraphics[width=.95\textwidth]{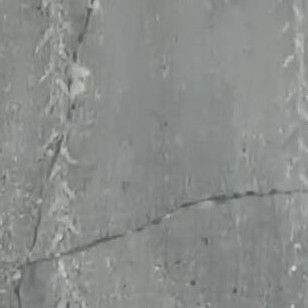}
    \end{subfigure}%
    \begin{subfigure}{.16\textwidth}
    \centering
    \includegraphics[width=.95\textwidth]{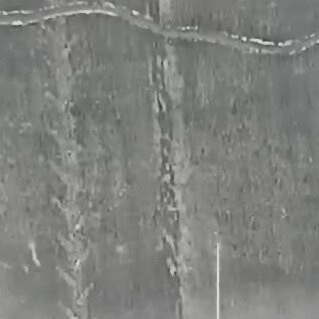}
    \end{subfigure}
    
    \vspace{1ex}
    \begin{subfigure}{.16\textwidth}
    \centering
    \includegraphics[width=.95\textwidth]{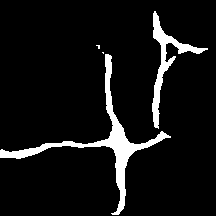}
    \end{subfigure}%
    \begin{subfigure}{.16\textwidth}
    \centering
    \includegraphics[width=.95\textwidth]{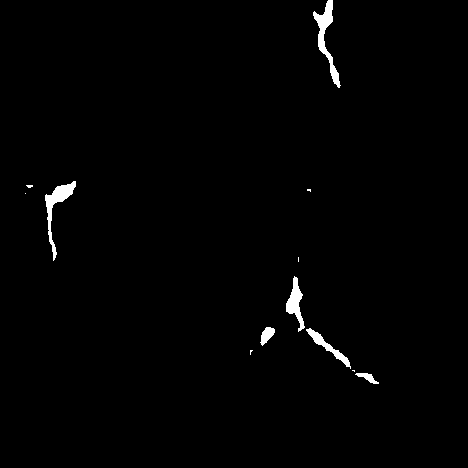}
    \end{subfigure}%
    \begin{subfigure}{.16\textwidth}
    \centering
    \includegraphics[width=.95\textwidth]{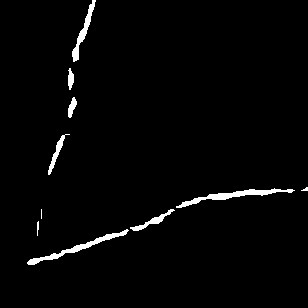}
    \end{subfigure}%
    \begin{subfigure}{.16\textwidth}
    \centering
    \includegraphics[width=.95\textwidth]{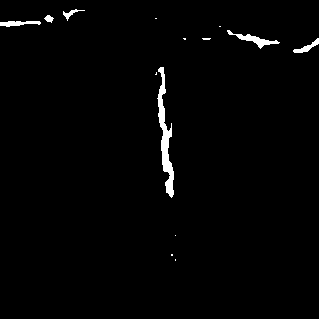}
    \end{subfigure}
    
    \vspace{1ex}
    \begin{subfigure}{.16\textwidth}
    \centering
    \includegraphics[width=.95\textwidth]{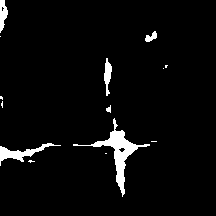}
    \end{subfigure}%
    \begin{subfigure}{.16\textwidth}
    \centering
    \includegraphics[width=.95\textwidth]{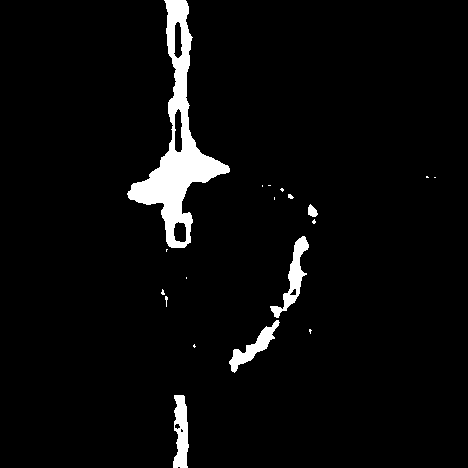}
    \end{subfigure}%
    \begin{subfigure}{.16\textwidth}
    \centering
    \includegraphics[width=.95\textwidth]{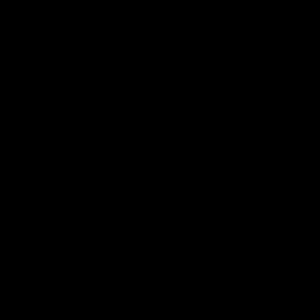}
    \end{subfigure}%
    \begin{subfigure}{.16\textwidth}
    \centering
    \includegraphics[width=.95\textwidth]{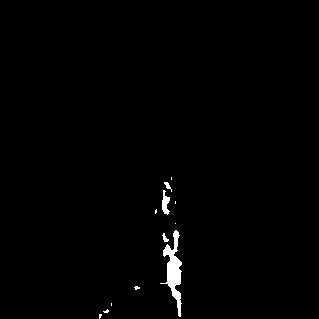}
    \end{subfigure}
    
    \vspace{1ex}
    \begin{subfigure}{.16\textwidth}
    \centering
    \includegraphics[width=.95\textwidth]{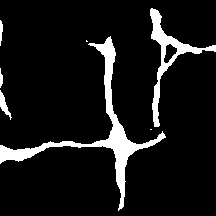}
    \end{subfigure}%
    \begin{subfigure}{.16\textwidth}
    \centering
    \includegraphics[width=.95\textwidth]{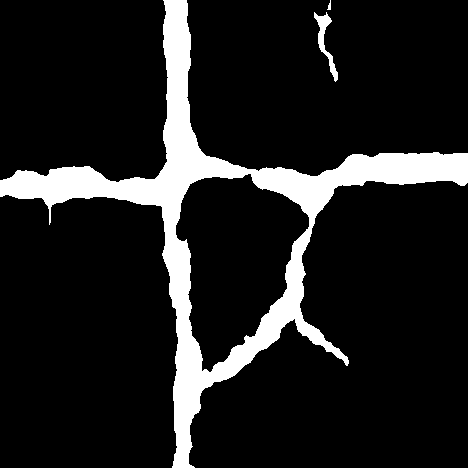}
    \end{subfigure}%
    \begin{subfigure}{.16\textwidth}
    \centering
    \includegraphics[width=.95\textwidth]{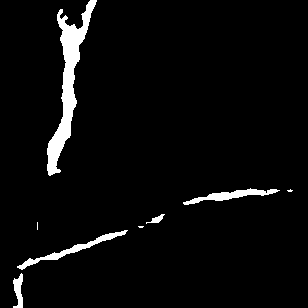}
    \end{subfigure}%
    \begin{subfigure}{.16\textwidth}
    \centering
    \includegraphics[width=.95\textwidth]{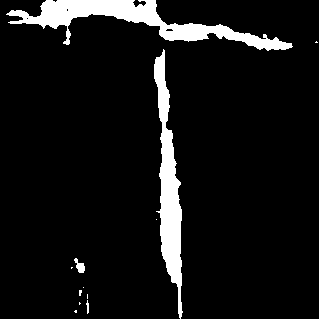}
    \end{subfigure}
    
    \vspace{1ex}
    \caption{Visualization of the airport pavement defect detection results. The first row is the input image. The other rows are the results of Experiments 4, 5 and 6 respectively (see Table~\ref{tab:runwaymetrics}).}
    \label{fig:prediction_examples}
\end{figure}

Some visual examples of the results are shown in Figure~\ref{fig:prediction_examples}.

When checking the results visually, overall the model trained with a mixed dataset performs the best in every experiment. For example, in column 2, where there is a crack with grass filled in, the model trained with real images is not able to detect it at all, whereas the other two models can detect it with different accuracy. However, in the third column, the model trained with synthetic images predicts a blank image with no defects, whereas the models trained using real data can correctly predict the crack's position.

With this subjective analysis, we conclude that the segmentation model performs well, despite the relatively low metric values. This could be attributed to the pixel-level difference between the annotation and the model output. It is perceptible in the sample images that most of the cracks are identified properly, but non-accurate annotations will output lower metric values.

\section{Conclusion}
In this work, we have presented a methodology for creating synthetic images for training a pavement defect segmentation method. We provide comprehensive steps for scenario generation using a 3D graphics engine and drone simulation software, to capturing the images with their corresponding ground truth.

We also designed a deep learning model for crack segmentation, which performs real-time prediction in an embedded computer. This model performs similarly to other state-of-the-art crack segmentation methods with relatively low computational requirements.

We found that adding synthetic data to the training set can improve prediction accuracy when the available datasets are of a slightly different domain and demonstrated this claim with the crack segmentation use case in runway inspection, where we saw a significant improvement when training with a mixed real and synthetic dataset, compared to real images alone.

To further improve the segmentation results, in future works domain adaptation techniques could be used, which would enable better transfer between virtual and real scenarios. The synthetic data could also be diversified by adding more crack variations or with methods to generate crack shapes and place them into the virtual environment.

\acknowledgments
This work has been partially supported by project 5D-Aerosafe funded by the European Union's H2020 research and innovation programme (grant agreement 861635).

\bibliography{egbib} 
\bibliographystyle{spiebib} 
\end{document}